%% file: main.tex
\documentclass[final]{conference}

\usepackage{times}
\usepackage{epsfig}
\usepackage{graphicx}
\usepackage{amsmath}
\usepackage{amssymb}

\usepackage{graphicx}
\usepackage{amsmath,amssymb} 
\usepackage{color}
\usepackage[ruled,vlined]{algorithm2e}
\usepackage[british,UKenglish,USenglish,american]{babel}
\usepackage{comment}
\usepackage{multirow}
\usepackage{makecell}
\usepackage{subcaption}
\usepackage{cancel}
\usepackage{bm}
\usepackage{mathrsfs}
\usepackage{physics} 

\DeclareMathOperator*{\argmax}{arg\,max} 
\DeclareMathOperator*{\sigmoid}{sigmoid} 

\usepackage[pagebackref=true,breaklinks=true,colorlinks,bookmarks=false]{hyperref}

\begin{document}

\title{Unsupervised Object Detection with LiDAR Clues}

\author{Hao Tian$^{1*}$, Yuntao Chen$^{2,3}$\thanks{Equal contribution. $^{\dag}$ This work is done when Hao Tian and Yuntao Chen are interns at SenseTime Research.}~~$^{\dag}$, Jifeng Dai$^{1}$, Zhaoxiang Zhang$^{2,3,4}$, Xizhou Zhu$^{1}$  \\
$^{1}$SenseTime Research \quad 
$^{2}$University of Chinese Academy of Sciences \\
$^{3}$Center for Research on Intelligent Perception and Computing, CASIA\\
$^{4}$Center for Excellence in Brain Science and Intelligence Technology, CAS\\
\texttt{\{zhuwalter,tianhao1,daijifeng\}@sensetime.com} \\
\texttt{\{chenyuntao2016,zhaoxiang.zhang\}@ia.ac.cn}\\
}

\maketitle
\input{srcs/0-abstract.tex}

\input{srcs/1-introduction.tex}

\input{srcs/2-related.tex}

\input{srcs/3-method.tex}

\input{srcs/4-experiment.tex}

\input{srcs/5-conclusion.tex}

\input{srcs/7-acknowledgement.tex}

{\small
\bibliographystyle{ieee_fullname}
\bibliography{egbib}
}

\input{srcs/6-appendix.tex}

\end{document}

%% file: srcs/0-abstract.tex
\begin{abstract}

Despite the importance of unsupervised object detection, to the best of our knowledge, there is no previous work addressing this problem.
One main issue, widely known to the community, is that object boundaries derived only from 2D image appearance are ambiguous and unreliable. To address this, we exploit LiDAR clues to aid unsupervised object detection.  By exploiting the 3D scene structure, the issue of localization can be considerably mitigated. We further identify another major issue, seldom noticed by the community, that the long-tailed and open-ended (sub-)category distribution should be accommodated. In this paper, we present the first practical method for unsupervised object detection with the aid of LiDAR clues. In our approach, candidate object segments based on 3D point clouds are firstly generated.
Then, an iterative segment labeling process is conducted to assign segment labels and to train a segment labeling network, which is based on features from both 2D images and 3D point clouds. The labeling process is carefully designed so as to mitigate the issue of long-tailed and open-ended distribution.
The final segment labels are set as pseudo annotations for object detection network training.
Extensive experiments on the large-scale Waymo Open dataset suggest that the derived unsupervised object detection method achieves reasonable accuracy compared with that of strong supervision within the LiDAR visible range.

\end{abstract}

%% file: srcs/1-introduction.tex
\section{Introduction}
\label{sec:intro}

\begin{figure}[t]
    \centering
    \includegraphics[width=0.9\linewidth]{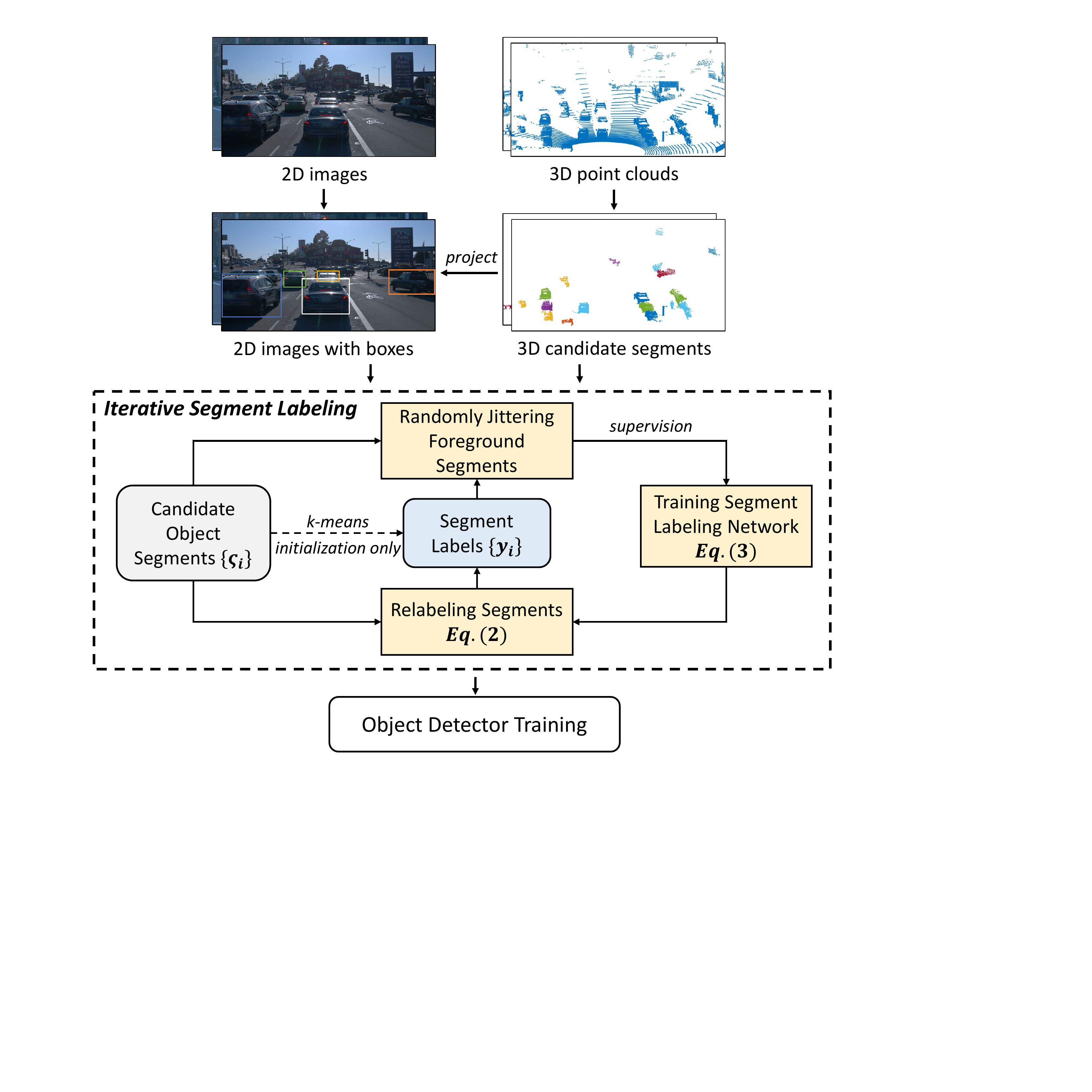}
    \vspace{-0.5em}
    \caption{Illustration of the proposed approach.}
    \label{fig:algo_overview}
    \vspace{-1.5em}
\end{figure}

\begin{figure*}
  \begin{subfigure}[t]{0.5\linewidth}
    \centering
    \includegraphics[width=0.9\linewidth]{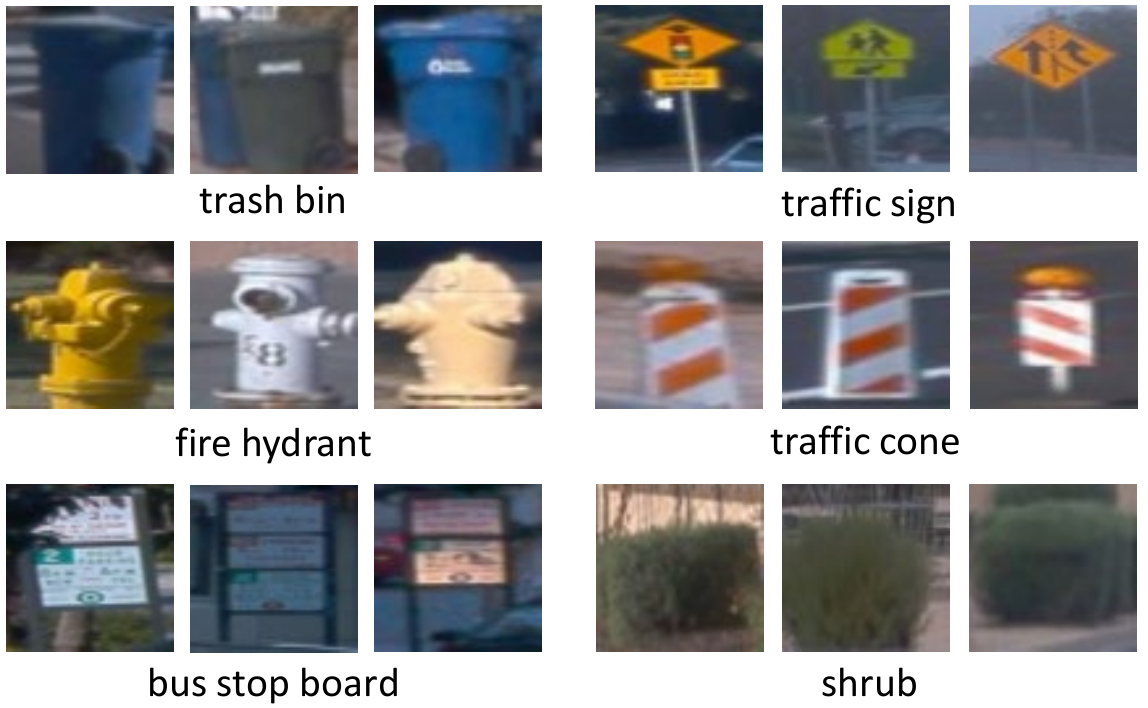}
    \caption{unlabeled categories.}
    \label{fig:clusters}
  \end{subfigure}
  \begin{subfigure}[t]{0.5\linewidth}
    \centering
    \includegraphics[width=0.9\linewidth]{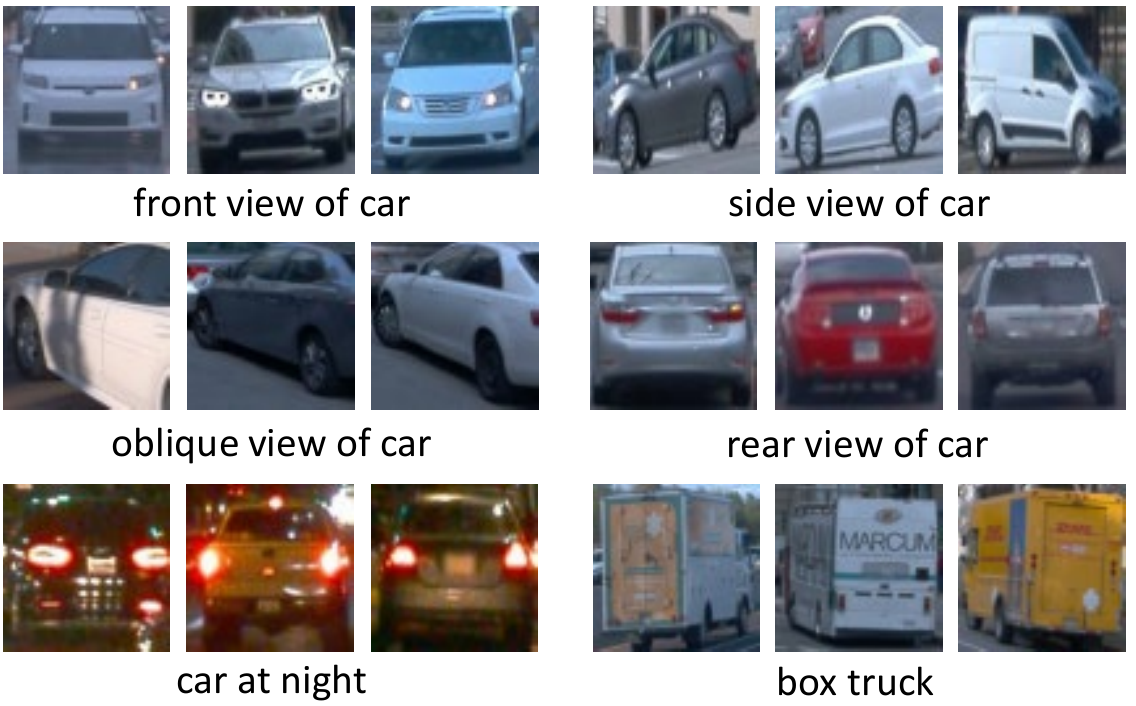}
    \caption{semantic sub-groups in the `vehicles' category.}
    \label{fig:subgroup}
  \end{subfigure}
  \vspace{-0.5em}
  \caption{Examples of object categories and semantic sub-groups discovered in the training images, which are not annotated in the Waymo Open dataset~\cite{sun2019scalability}.}
  \label{fig:vis_cat}
  \vspace{-1em}
\end{figure*}

Unsupervised object detection requires localization and classification of object instances without manual annotations in 2D images.
Due to the importance of the problem, various relevant tasks have been studied. For example, some weakly-supervised object detection methods~\cite{cinbis2016weakly,diba2017weakly,bilen2016weakly,kantorov2016contextlocnet,tang2017multiple} seek to detect objects with image-level annotations only, while some semi-supervised object detection methods~\cite{radosavovic2018data,jeong2019consistency,tang2020proposal} are trained on both bounding box annotated data and additional massive unlabeled images. Unsupervised object proposal generation~\cite{carreira2010constrained,endres2010category,uijlings2013selective,ren2013image,arbelaez2014multiscale} has also been widely studied. However, to the best of our knowledge, there is no previous work addressing the unsupervised object detection problem.

Recently, considerable progress has been made in unsupervised feature learning~\cite{liu2020self}. The networks with the unsupervised learned features achieve accuracies on par with those of strong supervision when fine-tuned on down-stream tasks. In this trend, some cluster discrimination based methods~\cite{xie2016unsupervised,ji2019invariant,haeusser2018associative,caron2018deep,yang2016joint,chang2017deep,van2020scan} have tried to address the \textit{unsupervised image classification} problem. Competitive results compared with semi-supervised learning on ImageNet~\cite{deng2009imagenet} are obtained in \cite{van2020scan}.
However, there is still a significant gap between unsupervised classification and unsupervised object detection, which involves both localizing and classifying multiple object instances in images.

One widely-known issue for unsupervised object detection is how to localize object instances precisely from the cluttered background without any human annotations. Unsupervised object localization from 2D images is extremely challenging because objects are of heterogeneous colors and textures with various shapes and occlusions. For example, unsupervised object proposals are usually generated by merging over-segmented regions according to color or texture clues~\cite{carreira2010constrained,endres2010category,uijlings2013selective,ren2013image,arbelaez2014multiscale}, which suffer from ambiguous and unreliable object boundaries. Although the object recall rate can be satisfactory with large proposal numbers, the precision rate is very low due to the ambiguities.

For the localization challenge, we argue that the key missing piece is the 3D scene structure, which is essential for human vision. Object boundaries can be better distinguished in the 3D point cloud because different objects cannot occupy the same 3D location. On the other hand, 3D shape information can also be used to better classify objects. Thanks to the popularization of LiDAR sensors, such synchronized 2D images and 3D point clouds have become much easier to obtain. Here we choose to generate candidate object segments based on 3D topology and to learn to label these segments into different categories / clusters\footnote{The predictions made by our approach are actually of clusters because no annotations are utilized. But for the coherence of terminology, we use the term categories without confusion in the paper.}. The labeling predictions are also based on features of both 2D images and 3D point clouds.

With the localization issue mitigated, we further identify another major issue, seldom noticed by the community: accommodating the long-tailed and open-ended (sub-)category distribution in unsupervised object detection. Research works in unsupervised classification~\cite{xie2016unsupervised,ji2019invariant,haeusser2018associative,caron2018deep,yang2016joint,chang2017deep,van2020scan} are mostly experimented on balanced and closed-world datasets (e.g., ImageNet~\cite{deng2009imagenet} and CIFAR~\cite{krizhevsky2009learning}), which consist of known number of categories with balanced number of images. In unsupervised object proposal generation~\cite{carreira2010constrained,endres2010category,uijlings2013selective,ren2013image,arbelaez2014multiscale}, all the object categories merge into one single class of foreground objects. Thus, the long-tail and open-ended distribution problem does not bother. However, for object detection, the datasets (e.g., LVIS~\cite{gupta2019lvis} and Open Images~\cite{kuznetsova2018open}) often have a long-tailed distribution, due to the nature of natural scenes. In the situation of unsupervised object detection, the difficulty is even further magnified, where the actual number of appeared object categories is unknown. Besides, the long-tailed and open-ended distribution not only exists in object categories but also exists in different semantic sub-groups (e.g., different views, poses) of the same object category. There is no hint which sub-groups belong to the same category till final evaluation or human examination. These semantic sub-groups need to be accurately discovered and detected so as to capture the whole semantic categories. The challenge of long-tailed and open-ended distribution brings difficulty for the labeling of candidate object segments. The head categories contain many object segments, while the tail categories contain very few. Proper labeling mechanism is necessary so as to avoid the tail categories being buried with the cluttered background.

This paper presents the first practical method for unsupervised object detection with the aid of LiDAR clues. The input is a training set composed of synchronized 2D image and 3D point cloud pairs without any type of human annotations, while the output is an object detection network applicable to 2D images. Our approach is illustrated in Figure~\ref{fig:algo_overview}. For each training pair, candidate object segments are first extracted from the 3D point cloud, based on the 3D topology instead of 2D image appearance. Iterative segment labeling is then conducted to assign segment labels and to train a segment labeling network, assuming segments with similar 2D image appearances and 3D shapes are of the same category. Such iterative optimization makes the predicted categories fit the long-tailed and open-ended distributions. The final segment labels are set as pseudo annotations for object detection network training. During testing, the trained detection network is applied to 2D images.

In iterative labeling, the segment labeling network and its training mechanism are carefully designed to accommodate the nature of the long-tailed and open-ended distribution. In the segment labeling network, motivated by~\cite{tan2020equalization}, we cancel the competition among foreground categories to prevent misclassifying objects in tail categories as background. During network training, only segments labeled as foreground are utilized to provide training losses, to avoid the impact of those foreground segments wrongly labeled as background. The negative examples are generated by jittering the segments labeled as foreground. Starting from a large number of allowed object categories (10,000 by default), our approach will automatically discover the effective number and distribution of appeared categories.

Extensive experiments on the large-scale Waymo Open dataset~\cite{sun2019scalability} suggest that the derived unsupervised object detection method achieves reasonable accuracy compared with that of strong supervision within the LiDAR visible range.
Besides, our approach can detect object categories appear in the training images but are not annotated in the dataset, such as `trash bin', `traffic sign', and `fire hydrant'. Figure~\ref{fig:vis_cat} and Figure~\ref{fig:det_results} show some example results of our proposed approach. Code shall be released.

%% file: srcs/2-related.tex
\section{Related Work}

\noindent\textbf{Discriminative Unsupervised Feature Learning~} 
Recently, learning visual features by discriminative tasks without human supervision has shown great promise~\cite{liu2020self}. The networks with the learned features achieve accuracy on par with those of strong supervision when fine-tuned on down-stream tasks. Most approaches can be categorized into two classes: instance discrimination or cluster discrimination.

\textit{Instance discrimination} methods treat each image in a dataset as an individual instance and learn discrimination among instances. \cite{dosovitskiy2015discriminative} firstly proposed to learn a classifier with each image as a category. InstDisc~\cite{wu2018unsupervised} replaces the classifier with a contrastive loss over the memory bank that stores previously-computed representations for other images. CMC~\cite{tian2019contrastive} further extends it by taking multiview of the same image as positive samples. MoCo~\cite{he2020momentum,chen2020improved} improves the contrastive learning methods by storing representations from a momentum encoder. SimCLR~\cite{chen2020simple,chen2020big} shows that the memory bank is unnecessary if the batch size is large enough. Very recently, BYOL~\cite{grill2020bootstrap} even discards
negative sampling in self-supervised learning.

\textit{Cluster discrimination} methods~\cite{caron2018deep,caron2019unsupervised,zhuang2019local,asano2019self,huang2019unsupervised,gidaris2020learning,caron2020unsupervised} learn discrimination among groups of images with similar appearance instead of among individual images. In DeepCluster~\cite{caron2018deep}, given the encoded features, k-means is applied to generate pseudo labels. The encoded features are further refined by learning to classify according to these pseudo labels. \cite{caron2019unsupervised} scales this method to massive non-curated data. \cite{zhuang2019local} improves the performance by replacing the mutual-exclusive clustering with a local soft-clustering. SeLa~\cite{asano2019self} and SwAV~\cite{caron2020unsupervised} further formulate the problem as simultaneous clustering and representation learning by maximizing the information between pseudo labels and input data indices.

Most of these methods focus on the quality of the learned features, where the performance is usually evaluated via linear classifier with fixed features, few-shot image classification, and transfer learning~\cite{goyal2019scaling}. While some cluster discrimination methods~\cite{xie2016unsupervised,ji2019invariant,haeusser2018associative,caron2018deep,yang2016joint,chang2017deep,van2020scan} would also directly evaluate the quality of clustering. By viewing the task as \textit{unsupervised classification}, they evaluate how well the learned clusters align with the semantic categories. Typically, Normalized Mutual Information~\cite{caron2018deep,yang2016joint,chang2017deep,van2020scan}, Adjusted Rand Index~\cite{chang2017deep,van2020scan}, or Accuracy under the best mapping between learned clusters and ground-truth categories~\cite{xie2016unsupervised,ji2019invariant,haeusser2018associative,chang2017deep,van2020scan} are used as the evaluation metrics.

The cluster discrimination methods are most relevant to the proposed method. In our approach, we also iteratively refine the cluster assignment and conduct feature learning. The key difference is we leap forward from image classification to object detection, which involves both localizing and classifying multiple object instances. Besides, the datasets in image classification are usually balanced and of closed-world. While we conduct unsupervised object detection on long-tailed and open-world distributions.

\vspace{0.5em}
\noindent\textbf{Unsupervised-, Weakly- and Semi-Supervised Object Detection~} 
To reduce the hunger of human annotations, various training settings have been studied for object detection.~\cite{cinbis2016weakly,diba2017weakly,bilen2016weakly,kantorov2016contextlocnet,tang2017multiple} seek to detect objects with image-level annotations only. 
While \cite{hoffman2014lsda,tang2016large,redmon2017yolo9000} train object detectors on data with bounding box annotations for some categories and image-level annotations for other categories. Another prevalent setting is training on bounding box annotated data with additional massive unlabeled data~\cite{radosavovic2018data,jeong2019consistency,tang2020proposal}. Recently, there are also some works focusing on training with partial bounding box annotated images~\cite{wu2018soft,xu2019missing,yoon2020semi,zhang2020solving}. 

There are also some relevant tasks focusing on unsupervised object localization, where the classification of object instances is not involved.
\textit{Unsupervised object proposal generation} has been widely studied on static 2D images (e.g., by grouping similar super-pixels)~\cite{carreira2010constrained,endres2010category,uijlings2013selective,ren2013image,arbelaez2014multiscale}. There are also research works on unsupervised proposal generation on videos~\cite{sharir2012video,van2013online,oneata2014spatio,fragkiadaki2015learning,xiao2016track}, 3D point clouds~\cite{karpathy2013object,collet2013exploiting,douillard2011segmentation,bogoslavskyi2016fast,zermas2017fast,bogoslavskyi2017efficient,hasecke2020fast}, etc.
\textit{Motion segmentation} is a binary labeling task of identifying the individual moving pixels in videos w.r.t. the background motion~\cite{bovik2009essential}, which is conducted on 2D image frames via background modeling~\cite{papazoglou2013fast,yang2019unsupervised} or structure from motion~\cite{namdev2012motion,yin2018geonet,ranjan2019competitive,ranftl2016dense}. 

To the best of our knowledge, there is no previous work addressing the unsupervised object detection problem. Compared with weakly- and semi-supervised object detection methods, we do not use any type of human annotation. Our approach is relevant to unsupervised object localization. Here we utilize the unsupervised object proposal generation approach as a component in our method.

\vspace{0.5em}
\noindent\textbf{Supervised 3D Object Detection~}
A typical 3D object detector takes the point cloud of a scene as its input and produces oriented 3D bounding boxes localizing each detected object~\cite{guo2020deep}. These methods can be divided into two categories: region-proposal-based and single-shot-based methods. Region-proposal-based methods generate and classify the region proposals, which typically use multi-view~\cite{chen2017multi,ku2018joint,liang2018deep,liang2019multi}, point cloud segmentation~\cite{yang2019std,shi2019pointrcnn,yang2018ipod} or frustum~\cite{wang2019frustum,xu2018pointfusion,qi2018frustum,shin2019roarnet} as representations.
Single-shot-based methods directly predict class probabilities and regress 3D bounding boxes via single-stage networks. The network is applied on Bird's Eye View (BEV)~\cite{beltran2018birdnet,yang2018hdnet,yang2018pixor}, discretized voxel~\cite{li2016vehicle,engelcke2017vote3deep,li20173d,zhou2018voxelnet,yan2018second,lang2019pointpillars}, or point clouds~\cite{yang20203dssd} as representations. 

Here we also use point cloud segmentation to generate region proposals for object detection. Aside from being unsupervised, the key difference is our approach utilizes the point clouds in train-time only, providing 3D information to aid unsupervised object detector training. Our trained object detectors are applied to 2D images.

%% file: srcs/3-method.tex
\section{Algorithm}

\subsection{Overview}

In our approach, the input is a training set composed of 2D image and 3D point cloud pairs, capturing natural scenes at synchronized times. Note that no annotations are available in the training set. The output is an object detection network applicable to 2D images, which is capable of detecting object categories seen in training\footnote{The trained object detection network can localize and classify the objects in the test input image. However, it is unaware of the semantic naming of the categories. For example, the network can localize frontal and side-view cars in the image, and classify them to be of ``1-st category" and ``2-nd category" respectively. But it has no information to associate the ``1-st category" and the ``2-nd category" to ``car". Further category naming association is necessary for numerical evaluation, as described in Section~\ref{sec:detector}.}.

Figure~\ref{fig:algo_overview} provides an overview of our approach.
For each training pair, candidate object segments are extracted from the 3D point clouds, based on the 3D topology. Each object segment is represented by its corresponding 3D points. A segment labeling network is trained to assign labels for the candidate segments, indicating whether the segments highly overlap with the object instances (foreground / background), and which categories they belong to. Iterative optimization is conducted to assign labels and to train the segment labeling network. Assuming segments with similar 2D image appearances and 3D shapes are of the same category, we may expect such iterative optimization can correct inconsistent segment labels. The final segment labels are set as pseudo annotations for object detection network training. The trained network is applied to 2D images.

\subsection{Exploiting 2D Images and 3D Point Clouds}

\vspace{0.5em}
\noindent\textbf{Object Segment Generation on Point Cloud~} We generate a set of candidate segments based on some object proposal approaches. Foreground objects and their object categories are discovered by labeling the generated object segments. We found it more reliable to differentiate objects on 3D point clouds than on 2D images because different objects cannot occupy the same 3D location. Here we utilize the 3D point cloud segmentation algorithm in~\cite{bogoslavskyi2017efficient} to generate candidate segments. For each image $x$ and its corresponding 3D point cloud $P$, the generated candidate segments are represented by $S =  \{\varsigma_i\}_{i=1}^{n}$,  where $n$ is the number of segments, and $\varsigma_i$ denotes the $i$-th segment. A segment $\varsigma_i \subseteq P$ is a collection of 3D points, as a subset of point cloud $P$. Each segment $\varsigma_i$ corresponds to a 2D bounding box $b_i$, by projecting it on the image plane. Some segments may well localize the objects, while others are on the cluttered background.

\vspace{0.5em}
\noindent\textbf{Exploiting 2D Images and 3D Point Clouds in Segment Labeling Network~}  
The segment labeling network $N$ takes both the 2D image and 3D point cloud as input. There are two sub-network extracting 2D image features and 3D point cloud features, respectively. 
Given the 3D segment $\varsigma_i$, PointNet~\cite{qi2017pointnet} is applied to extract a 1024-d 3D shape feature. Given the image $x$ and the 2D bounding box $b_i$, Fast R-CNN~\cite{girshick2015fast} with ResNet-50~\cite{he2016deep} is applied to extract a 1024-d 2D image appearance feature for the $i$-th segment. The concatenation of these two features is further fed into a linear classifier, producing a C-d vector ($C$ denotes the number of foreground categories) followed by a sigmoid function to predict the foreground category probabilities $\bm{s}_i$, which shall be further discussed in network design \& training in Section~\ref{sec:optimization}.

\subsection{Iterative Segment Labeling}
\label{sec:optimization}

Assuming the maximum number of allowed object categories is $C$,
for each image $x$, labels $Y = \{y_i\}_{i=1}^{n}$ are assigned for all the candidate segments $S =  \{\varsigma_i\}_{i=1}^{n}$, where $y_i \in \{0, 1, ..., C\}$. $y_i = 0$ indicates the segment $\varsigma_i$ is on cluttered background. $y_i = c > 0$ indicates the segment $\varsigma_i$ highly overlaps with an object belonging to the $c$-th category. In unsupervised object detection, the (sub-)category distribution is long-tail and open-ended. Because there is no hint whether a tail category corresponds to / is a sub-group for an annotated category for evaluation, we need to ensure the labeling quality for both the head and tail categories. Here the segment labeling is conducted in an iterative manner, where the assigned labels gradually conform to the underlying long-tail and open-ended distribution.

\vspace{0.5em}
\noindent\textbf{Segment Label Initialization~} First, we need to derive an initial guess of the labels $\{y_i\}_{i=1}^{n}$. Here each object segment is assigned as one of the $C$ foreground categories. We conduct clustering based on the concatenation of the PointNet features extracted on 3D segment $\varsigma_i$, and the ResNet-50 features extracted from 2D image patch within bounding box $b_i$. The PointNet sub-network is randomly initialized. While the ResNet-50 backbone is initialized with ImageNet~\cite{deng2009imagenet} MoCo v2~\cite{chen2020improved} pre-trained weights. Note that no annotated labels are involved here. The 3072-d concatenation of PointNet and ResNet-50 features form the representation of the object segments. The features are tuned by applying MoCo v2 self-supervised training, where each object segment is treated as an individual sample. The PointNet and ResNet-50 network weights are updated accordingly. After MoCo v2 training, k-means
clustering~\cite{bahmani2012scalable} is performed on the trained  3072-d object segment features (with $C$ clusters). Each object segment is assigned with the corresponding cluster index as its initial category label, $y_i \in \{1, 2, ..., C\}$.

\vspace{0.5em}
\noindent\textbf{Segment Labeling Network Design and Training~} At each round in the iterative segment labeling, given the segment labels $\{y_i\}_{i=1}^{n}$ generated from the previous round, the segment labeling network is re-trained from initialization. Note that no annotated data is utilized in initializing the segment labeling network. The PointNet sub-network is initialized from the weights in segment label initialization. While the ResNet-50 backbone in Fast R-CNN is initialized with ImageNet MoCo v2 pre-trained weights. The remaining weights in the network are randomly initialized.

Due to the long-tail distribution, a straight-forward implementation of letting segment labeling network directly predict $y_i$ does not work well. The segments corresponding to tail categories are usually low scored, which are hard to be differentiated from the background. Careful sample reweighing in training the segment labeling network may mitigate the problem. But in our approach, the sample labels are actually the predictions by the segment labeling network in the previous iteration. The errors can get magnified if the sample reweighing is not well-tuned. 

Here, we find a simple design of the segment labeling network can make it much more robust to long-tail distribution. Motivated by~\cite{tan2020equalization}, in the segment labeling network, the probabilities of each foreground category are independently predicted using the sigmoid function. In the design, no competition is involved between foreground categories. Each foreground category only needs to differentiate itself from the background clutters, as
\begin{equation}
\bm{s}_i = \sigmoid\left( N(x, b_i, \varsigma_i | \theta) \right),
\end{equation}
where $N$ denotes the object segment classification network, $\theta$ is the network parameter, $\bm{s}_i \in [0, 1]^C$ are the probabilities that segment $\varsigma_i$ belongs to different foreground object categories, which is trained with a simplified version of Equalization Loss~\cite{tan2020equalization}, which shall be further elaborated.

Given the scores $\bm{s}_i$ of $\varsigma_i$, the label $y_i$ is derived as
\begin{equation}
y_i = \begin{cases}
0,& \text{if } \max \bm{s}_i < \eta, \\
\argmax \bm{s}_i,& \text{otherwise},
\end{cases}
\end{equation}
where $\eta$ is the foreground probability threshold. If the segment $\varsigma_i$ has the estimated foreground probabilities $\bm{s}_i$ less than $\eta$ for all categories ($\eta = 0.95$ by default), it is assumed to be on background ($y_i = 0$). Otherwise, the category with the largest probability is set as the label ($y_i = \argmax \bm{s}_i$).

During training, object segments labeled as foreground are utilized to provide training losses. While the object segments labeled as background are discarded. This is because we find some actual foreground segments maybe wrongly labeled as background. Such labeling error deteriorates performance. Here we generate background training examples by randomly jittering the foreground segments. Given a foreground segment $\varsigma_i$ ($y_i > 0$) and its 2D bounding box $b_i$, random jittering is applied to produce jittered segment $\hat{\varsigma}_i$ and bounding box $\hat{b}_i$. Specifically, for each bounding box $b_i$, a target box IoU value $\text{IoU}_\text{target}$ is randomly sampled from a uniform distribution between 0.1 and 1.0. Then the jittered box $\hat{b}_i$ is generated by randomly sampling the top-left and bottom-right corners until the box IoU between $b_i$ and $\hat{b}_i$ is within $[\text{IoU}_\text{target}-0.005, \text{IoU}_\text{target}+0.005]$. The jittered segment $\hat{\varsigma}_i$ is derived from modifying $\varsigma_i$ according to $\hat{b}_i$. In it, the 3D points whose projected 2D coordinates locate outside of the jittered bounding box $\hat{b}_i$ are removed from $\varsigma_i$. The foreground / background label $\hat{z}_i \in \{0, 1\}$ is defined as $\hat{z}_i = 1$ if the jittered bounding box $\hat{b}_i$ overlaps with $b_i$ large than 0.5 in terms of box IoU, and $\hat{z}_i = 0$ otherwise.

These jittered segments serve as training samples. The training loss for each jittered segment is defined as
{\small
\begin{equation}
\label{eq:loss}
L(\hat{\bm{s}}_i; \hat{z}_i, y_i) = - \hat{z}_i \log(\hat{s}_{i, y_i}) - \sum_{c=1}^C (1 - \hat{z}_i) \log(1 - \hat{s}_{i, c})
\end{equation}}where $\hat{\bm{s}}_i = N(x, \hat{b}_i, \hat{\varsigma_i} | \theta)$ are the predictions made by the segment labeling network, $\hat{s}_{i, c}$ denotes the $c$-th value in $\hat{\bm{s}}_i$, which is the predicted probability that segment $\hat{\varsigma}_i$ belonging to the $c$-th category, $y_i$ is the segment label assigned in the previous iteration. 
This loss function is a simplified version of Equalization Loss~\cite{tan2020equalization}. In the Equalization Loss, to prevent misclassifying objects in tail categories as background, the classifier of each tail category will ignore the discouraging gradients from foreground samples of other categories. However, how to define tail categories needs careful tuning. We experimentally found that treating all categories as tail in the Equalization Loss works well. In our simplified Equalization Loss, jittered segments labeled as foreground ($\hat{z}_i = 1$) with category $y_i$ will encourage the probability prediction $\hat{s}_{i, y_i}$. Background segments  ($\hat{z}_i = 0$) will discourage the probability predictions for all foreground categories.
During training, the jittered segments with foreground and background labels are randomly sampled with a 1:3 ratio.

\vspace{0.5em}
\noindent\textbf{Summary~}
Figure~\ref{fig:algo_overview} illustrates the iterative segment labeling process.
First, we get the initial guess of the labels $\{y_i\}_{i=1}^{n}$ via k-means clustering~\cite{bahmani2012scalable} based on features learned with self-supervised training.
Then, the network training and segment labeling process are applied iteratively for several rounds. By such iterative labeling, the predicted categories fit the long-tailed and open-ended distribution.

Here, as the actual number of appeared object categories is unknown, the maximum number of allowed object categories $C$ is set as a large number in this paper (by default, $C=10,000$). We empirically observed that as the iteration runs, the produced segment labels naturally follow the long-tail distribution. The head categories occupy most candidate segments, while the tail categories are of the minority. Those non-existing categories within $\{1,\ldots,C\}$ finally do not claim any segments. More ablation studies are provided in Section~\ref{sec:ablation}.

\subsection{Object Detector Training and Evaluation~}
\label{sec:detector}

In our implementation, Faster R-CNN~\cite{ren2015faster} with FPN~\cite{lin2016feature} is chosen as the object detection network, where ImageNet MoCo v2 pre-trained ResNet-50 is utilized as the backbone. In training, the 2D bounding boxes and labeled categories of foreground segments ($\varsigma_i$ with $y_i \neq 0$) are set as pseudo annotations. During testing, the trained network is applied to 2D images.

A key point during training is to avoid feeding missed foreground objects as negative training examples.
In Faster R-CNN, for the RPN head, thanks to excessive background anchor boxes, by uniform random sampling (which is the default choice), the chances of sampling a missed foreground object as negative is low. But for the Fast R-CNN head, the negative region proposals produced by RPN have a good chance to be an actually missed foreground object. 
Therefore, we only collect region proposals with box IoU between 0.1 and 0.5 with the pseudo annotated boxes as negative examples for Fast R-CNN training. By forcing the sampled region proposals to overlap with the pseudo-annotated boxes, we considerably reduce the chance of sampling a missed foreground object as a negative example.

To mitigate the issue of long-tailed distribution, the simplified version of Equalization Loss~\cite{tan2020equalization} in Eq.~\eqref{eq:loss} is also applied for training the classifier in the Fast R-CNN head. Because the competition between foreground categories is canceled, each regional proposal may have multiple high-confidence category predictions. Class-agnostic NMS is utilized following the Fast R-CNN head to force each detected object with only one category label.

During the evaluation, we test the accuracy of the predictions made by the object detection network w.r.t. the annotated ground-truths. However, we have no idea which discovered cluster\footnote{Here we call the predictions by our approach \textit{cluster} to avoid confusion with the annotated categories.} in training corresponds to which semantic category annotated, without annotations or human examination. Besides, it is also possible that a discovered cluster captures just a subgroup of an annotated category. Thus, for quantitative evaluation, a many-to-one mapping between the discovered clusters and the ground-truth semantic categories is indispensable. Here we set up the mapping on the training set. Assuming there are $K$ ground-truth categories labeled in the dataset, each cluster in our approach is mapped to one of the $K$ labeled categories or a special ``others'' category. The ``others'' category is introduced because our approach will discover unlabeled categories. Specifically, for each generated candidate object segment during training, if the segment has overlapped with some ground-truth bounding boxes larger than 0.5, it will be assigned with the ground-truth category label of the bounding box with the largest overlap. Otherwise, the segment will be assigned as ``others''. Then, each cluster will be assigned to either a ground-truth category or the ``others'' category to achieve the minimum error rate. The category mapping and the object detector are finally applied on the test set, where the traditional AP metric can be utilized for evaluation.

%% file: srcs/4-experiment.tex
\section{Experiments}

\setcounter{dbltopnumber}{3}

\setlength{\tabcolsep}{3pt}
\setlength{\doublerulesep}{2\arrayrulewidth}
\renewcommand{\arraystretch}{1.2}
\begin{table*}[ht]
    \centering
    \small
    \resizebox{0.99\linewidth}{!}{
        \begin{tabular}{l|cc|c|cccc|cccc|cccc}
            \Xhline{2\arrayrulewidth}
            \multirow{2}{*}{annotation setting} & \multirow{2}{*}{\#images} & \multirow{2}{*}{\#bboxes} & network weights & \multicolumn{4}{c|}{vehicles} & \multicolumn{4}{c|}{pedestrians} &  \multicolumn{4}{c}{cyclists}\\
            \cline{5-16}
            & & & initialized from & AP & AP$_\text{S}$ & AP$_\text{M}$ & AP$_\text{L}$ & AP & AP$_\text{S}$ & AP$_\text{M}$ & AP$_\text{L}$ & AP & AP$_\text{S}$ & AP$_\text{M}$ & AP$_\text{L}$ \\
            \hline
            (a) our approach & 158k & 0 & ImageNet MoCo v2 & 28.7 & 1.0 & 34.0 & 77.2 & 28.5 & 6.8 & 46.0 & 60.2 & 1.0 & 0.0 & 0.0 & 1.7 \\
            \hline
            (b) manual annotations & 15k & 91k & ImageNet MoCo v2 & 29.2 & 1.0 & 33.8 & 83.2 & 22.7 & 1.9 & 38.9 & 57.0 & 3.9 & 0.0 & 3.3 & 17.9 \\
            (c) manual annotations & 15k & 91k & ImageNet supervised & 30.1 & 1.0 & 36.3 & 84.1 & 22.2 & 1.8 & 36.6 & 63.3 & 4.6 & 0.0 & 3.3 & 23.4 \\
            (d) manual annotations & 158k & 1087k & ImageNet MoCo v2 & 34.7 & 1.0 & 43.8 & 89.0 & 37.7 & 9.6 & 61.9 & 79.1 & 23.8 & 3.8 & 28.6 & 80.2 \\
            \hline
        \end{tabular}
    }
    \caption{Comparison of different annotation settings on Waymo Open validation.}
    \label{table:supervision_frcnn}
    \vspace{-0.3em}
\end{table*}

\setlength{\tabcolsep}{3pt}
\setlength{\doublerulesep}{2\arrayrulewidth}
\renewcommand{\arraystretch}{1.2}
\begin{table*}[t]
    \centering
    \small
    \resizebox{0.85\linewidth}{!}{
        \begin{tabular}{l|ccc|ccc|ccc}
            \Xhline{2\arrayrulewidth}
            \multirow{2}{*}{pre-training task and dataset}& \multicolumn{6}{c|}{Cityscapes} & \multicolumn{3}{c}{PASCAL VOC} \\
            \cline{2-10}
            & AP$^\text{bbox}$ & AP$^\text{bbox}_\text{50}$ & AP$^\text{bbox}_\text{75}$ & AP$^\text{mask}$ & AP$^\text{mask}_\text{50}$ & AP$^\text{mask}_\text{75}$ &  AP$^\text{bbox}$ & AP$^\text{bbox}_\text{50}$ & AP$^\text{bbox}_\text{75}$ \\
            \hline
            random initialization & 30.6 & 55.0 & 29.9 & 24.6 & 48.8 & 19.8 & 45.3 & 71.8 & 48.9 \\
            image classification on ImageNet & 37.5 & 62.1 & 38.5 & 31.8 & 56.9 & 29.7 & 54.6 & 80.9 & 60.8 \\
            our unsupervised object detection on Waymo & 36.1 & 60.8 & 36.2 & 30.6  & 57.1 & 27.0 & 53.8 & 79.3 & 59.3 \\
            \Xhline{2\arrayrulewidth}
        \end{tabular}
    }
    \caption{Transfer results of different pre-trained models on Cityscapes and PASCAL VOC.}
    \label{table:transfer}
    \vspace{-1em}
\end{table*}

\subsection{Dataset and Implementation Details}
\label{sec:implementation}

Evaluation is conducted on Waymo Open~\cite{sun2019scalability}, which is a recently released large-scale dataset for autonomous driving.
The dataset collects 2D videos and 3D point clouds at synchronized time steps.
The training and validation sets contain 798 and 202 videos with around 158k and 40k image frames, respectively. 
In experiments, 2D images from the `front' camera and 3D point clouds from the `top' LiDAR are utilized. The temporal information from videos is not used.
There are 3 object categories annotated with 2D bounding boxes in the dataset, i.e., `vehicles', `pedestrians', and `cyclists'. 
We test the feature transferability by fine-tuning on Cityscapes~\cite{Cordts2016Cityscapes} and PASCAL VOC~\cite{everingham2015pascal}. 

We also establish a manual-annotated detector baseline. 
In our approach, the discovered objects for training are all LiDAR visible. However, for the manual-annotated 2D bounding boxes, the LiDAR invisible distant boxes (with distance more than 75 meters) are also included. 
For a fair comparison, the manual-annotated baseline is trained with LiDAR-visible 2D bounding box annotations only.

Evaluation is measured by the traditional AP metric at the box IoU threshold of 0.5.
For better analysis, following COCO evaluation~\cite{lin2014microsoft}, results are also reported on small (area $<$ 32$^2$ pixels), medium (32$^2$ pixels $<$ area $<$ 96$^2$ pixels) and large objects (area $>$ 96$^2$ pixels), denoted as AP$_\text{S}$, AP$_\text{M}$ and AP$_\text{L}$, respectively.

In our approach, the iteration number for iterative segment labeling is set as 10 rounds when compared with manual annotations, and 1 round in ablation study by default, for experimental efficiency. Note that no manual annotation is involved until the final evaluation.
For all experiments, the hyper-parameter setting for Faster R-CNN~\cite{ren2015faster} follows the open source Detectron2~\cite{wu2019detectron2} code base.
For more dataset and implementation details please refer to Appendix.

\subsection{Comparison on Annotation Settings}

Table~\ref{table:supervision_frcnn} compares the results of our unsupervised object detection approach and those with manual annotations. 
Comparing Table~\ref{table:supervision_frcnn} (a) and (b), our purposed approach achieves on par accuracy on the `vehicles' and `cyclists' categories with that of using 10\% manual annotations, where a considerable higher AP is achieved on the `pedestrians' category.
Table~\ref{table:supervision_frcnn} (d) also presents the result of training with 100\% manual annotations, which achieves relatively high APs. Our unsupervised approach delivers reasonable accuracy compared to those of strongly supervised.

Because traditional supervised object detection methods use ImageNet supervised pre-training features, we also ablate the effect of unsupervised pre-training (i.e., MoCo v2). Comparing Table~\ref{table:supervision_frcnn} (b) and (c), the difference of unsupervised and supervised pre-training is negligible.

Table~\ref{table:transfer} presents the feature transferability performance by fine-tuning on Cityscapes~\cite{Cordts2016Cityscapes} and PASCAL VOC~\cite{everingham2015pascal}.
Pre-training by our unsupervised object detection approach demonstrates good feature transferability, which achieves on par performance with that of ImageNet pre-training, which is much better than that of random initialization.
Note that our approach only uses around 10\% number of unlabeled images for pre-training compared with that of ImageNet, which consists of 1.28M images with manual classification annotations. 

\subsection{Ablation Study}
\label{sec:ablation}

Due to limited space, here we only ablate some key factors. More detailed ablations are provided in Appendix. In the ablation studies, for experimental efficiency, only 1 round is applied in iterative segment labeling by default.

\setlength{\tabcolsep}{3pt}
\setlength{\doublerulesep}{2\arrayrulewidth}
\renewcommand{\arraystretch}{1.2}
\begin{table*}[t]
    \begin{subtable}[t]{0.40\textwidth}
        \centering
        \small
        \resizebox{0.99\linewidth}{!}{
        \begin{tabular}{lccc}
            \Xhline{2\arrayrulewidth}
            \multirow{2}{*}{setting} & \multicolumn{3}{c}{AP} \\
             & veh & ped & cyc \\
            \hline
            \makecell[l]{our approach} & 26.9 & 21.1 & 5.2 \\
            \hline
            \makecell[l]{-- segment labeling network $N$\\ \quad without point clouds} & 26.3 & 18.7 & 3.3 \\
            \hline
            \makecell[l]{--~-- candidate segments $S$ estimated \\ \quad~ by MCG~\cite{pont2016multiscale} on 2D images} & 11.0 & 0.0 & 0.0 \\
            \Xhline{2\arrayrulewidth}
        \end{tabular}}
        \caption{exploiting 2D images and 3D point clouds (1-st iter)}
        \label{table:components}
    \end{subtable}
    \hspace{\fill}
    \begin{subtable}[t]{0.25\textwidth}
        \centering
        \small
        \resizebox{0.95\linewidth}{!}{
        \begin{tabular}{cccc}
            \Xhline{2\arrayrulewidth}
            \multirow{2}{*}{\makecell[c]{loss \\ function}} & \multicolumn{3}{c}{AP} \\
             & veh & ped & cyc \\
            \hline
            softmax loss & 25.4 & 20.3 & 2.0 \\
            sigmoid loss & 23.8 & 20.7 & 0.6 \\
            \hline
            our approach & 28.7 & 28.5 & 1.0 \\
            \Xhline{2\arrayrulewidth}
        \end{tabular}}
        \caption{\parbox[t]{0.9\linewidth}{loss function for classification (10-th iter)}}
        \label{table:loss}
    \end{subtable}
    \hspace{\fill}
    \begin{subtable}[t]{0.27\textwidth}
        \centering
        \small
        \resizebox{0.79\linewidth}{!}{
        \begin{tabular}{cccc}
            \Xhline{2\arrayrulewidth}
            \multirow{2}{*}{\makecell[c]{video \\ sequences}} & \multicolumn{3}{c}{AP} \\
             & veh & ped & cyc \\
            \hline
            10\% & 23.1 & 15.5 & 1.6 \\
            25\% & 25.8 & 19.2 & 2.6 \\
            50\% & 26.0 & 21.6 & 4.9 \\
            \hline
            100\% & 26.9 & 21.1 & 5.2 \\
            \Xhline{2\arrayrulewidth}
        \end{tabular}}
        \caption{quantity of training data (1-st iter)}
        \label{table:quantity}
    \end{subtable}
    \vspace{-0.5em}
    \caption{Ablations on Waymo Open validation, where `veh', `ped' and `cyc' are abbreviations for `vehicles', `pedestrians' and `cyclists', respectively. Table~(a) gradually removes the usage of 3D point clouds. Table~(b) verifies our simplified version of Equalization Loss~\cite{tan2020equalization}. Table~(c) explores the effectiveness of training data.}
\end{table*}

\vspace{0.5em}
\noindent\textbf{Exploiting 2D Images and 3D Point Clouds~} 
During training, our unsupervised object detection approach utilizes 3D point clouds for object segment generation and in the segment labeling network.
Here, we ablate the necessity of 3D information.
Table~\ref{table:components} presents the results of not using 3D point clouds.
Without point clouds in the segment labeling network, the result is considerably worse, which indicates the classification can benefit from 3D shape information.
We further tried generating object segments on 2D images.
Here we adopt MCG~\cite{pont2016multiscale} to generate candidate segments from 2D images, which incurs large drop in AP.
This indicates  3D point clouds are important in localizing objects.

\vspace{0.5em}
\noindent\textbf{Iterative Segment Labeling~} 
Table~\ref{table:iterations} and Figure~\ref{fig:long_tail} present the results of applying different iteration rounds in the iterative segment labeling. At initialization, the number of object segments in different categories are close to uniform (see Figure~\ref{fig:long_tail}). As the algorithm iterates, the predicted categories gradually conform the long-tail distribution. The redundant categories in initialization disappear. The final detector accuracy also gets improved as the iteration runs, where the generated pseudo annotations fit better with the underlying distributions.

\begin{table*}[t]
\begin{minipage}[b]{0.32\textwidth}
      \centering
      \includegraphics[width=0.96\linewidth]{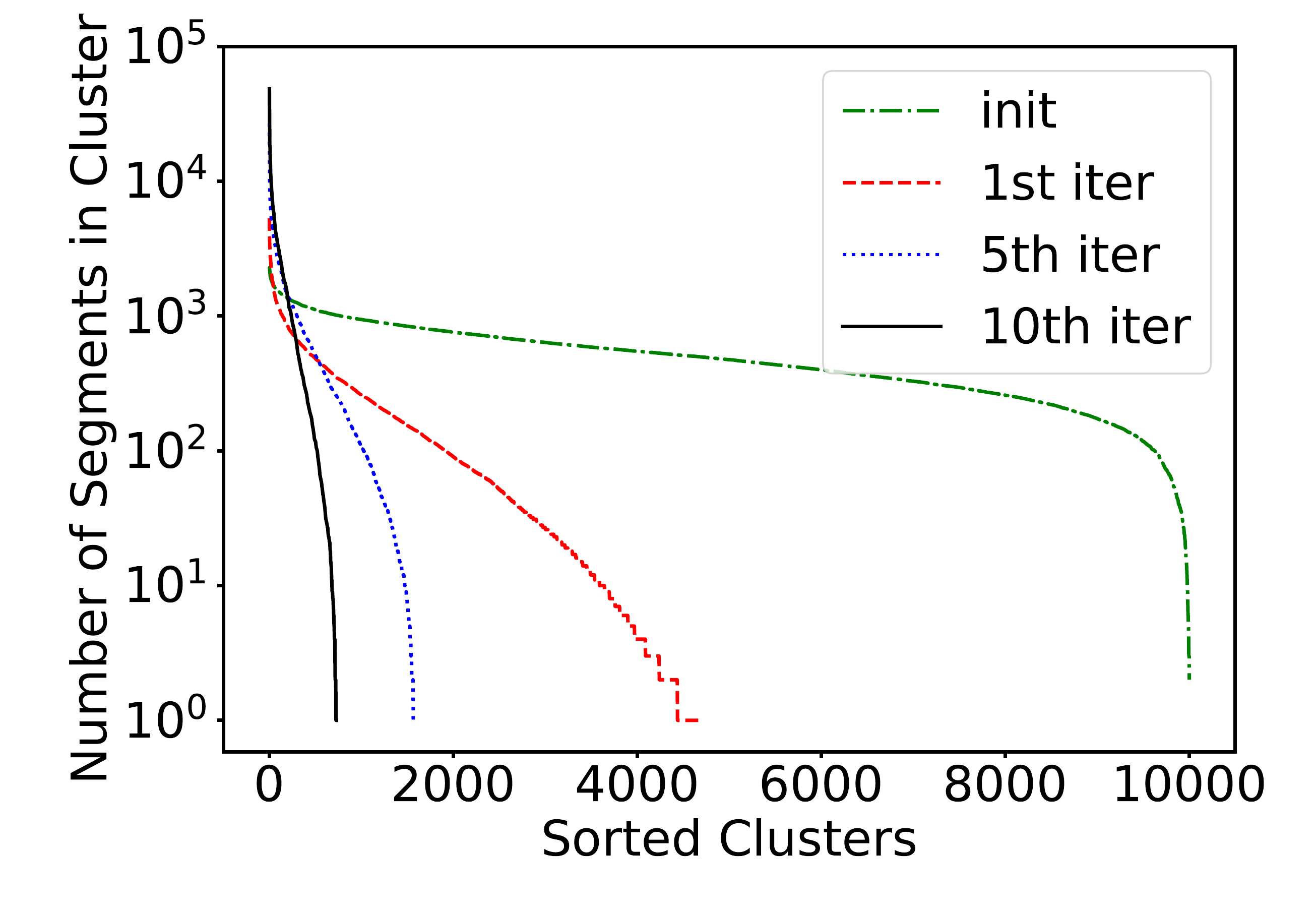}
      \vspace{-1.em}
      \captionof{figure}{Cluster distribution during the iterative segment labeling.}
      \label{fig:long_tail}
\end{minipage}
\hspace{\fill}
\begin{minipage}[b]{0.65\textwidth}
      \centering
      \small
      \resizebox{1.0\linewidth}{!}{
          \begin{tabular}{cc|cccccccccc}
              \Xhline{2\arrayrulewidth}
              \multicolumn{2}{c|}{iterations} & 1 & 2 & 3 & 4 & 5 & 6 & 7 & 8 & 9 & 10 \\
              \hline
              \multirow{3}{*}{AP} & vehicles & 26.9 & 27.7 & 28.5 & 28.2 & 27.9 & 28.7 & 28.7 & 28.9 & 28.8 & 28.7 \\
                                  & pedestrians & 21.1 & 23.8 & 25.6 & 26.8 & 27.0 & 27.9 & 28.0 & 28.3 & 28.5 & 28.5 \\
                                  & cyclists & 5.2 & 5.5 & 6.0 & 6.3 & 4.9 & 2.6 & 3.2 & 2.5 & 1.9 & 1.0 \\
              \hline
              \multirow{3}{*}{\#cluster} & 100\% segments & 4699 & 2921 & 2278 & 1889 & 1586 & 1353 & 1152 & 986 & 834 & 729 \\
                                         &  90\% segments & 1813 & 1282 & 982 & 772 & 610 & 489 & 402 & 332 & 281 & 243 \\
                                         &  80\% segments & 1240 & 888 & 673 & 521 & 403 & 320 & 260 & 216 & 185 & 161 \\
              \Xhline{2\arrayrulewidth}
          \end{tabular}
      }
      \caption{Ablation on iterations in the iterative segment labeling on Waymo Open validation.}
      \label{table:iterations}
\end{minipage}
\vspace{0.1em}
\end{table*}

\begin{figure*}[t]
    \centering
    \includegraphics[width=0.95\linewidth]{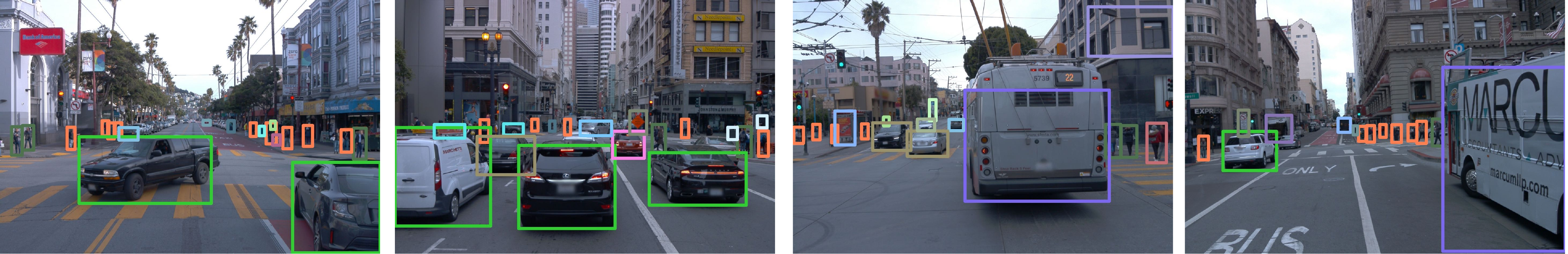}
    \caption{Object detection results on Waymo Open validation by Faster R-CNN trained with our unsupervised object detection method. Different colors of bounding boxes indicate the corresponding cluster id.}
    \label{fig:det_results}
    \vspace{-0.7em}
\end{figure*}

\vspace{0.5em}
\noindent\textbf{Loss Function for Foreground Classification~} In this work, we use a simplified version of Equalization Loss~\cite{tan2020equalization} (see Eq.~\eqref{eq:loss}) to mitigate the classification challenge of long-tailed distribution. Here we also tried using the vanilla softmax loss and sigmoid loss.
Table~\ref{table:loss} presents the results after 10 iterations. Using the vanilla softmax loss and sigmoid loss achieve relatively lower performance, which indicates the effectiveness of the proposed loss function.

\vspace{0.5em}
\noindent\textbf{Quantity of Training Data~}
We further explore the impact of data quantity. 
Different portions of video sequences are randomly sampled from Waymo for training. 
As showed in Table~\ref{table:quantity}, our approach can effectively exploit vast unlabeled images for unsupervised object detection.

\subsection{Visualization}
\label{sec:vis}
  
Example detection results produced by our unsupervised approach are presented in Figure~\ref{fig:det_results}.
Our approach can discover object categories and semantic subgroups in the training images but are not annotated.
Figure~\ref{fig:clusters} shows some discovered unlabeled categories. 
Figure~\ref{fig:subgroup} shows some discovered semantic subgroups in the `vehicles' category.

%% file: srcs/5-conclusion.tex
\section{Conclusion}

In this paper, we present the first practical method for unsupervised object detection.
During training, 3D point clouds are utilized to mitigate the difficulty of differentiating and localizing objects.
We further identify another major issue, seldom noticed by the community: the accommodation of  the long-tailed and open-ended distribution in object (sub-)categories. A carefully designed iterative segment labeling process is conducted to generate pseudo annotations for object detection network training. Extensive experiments on the large-scale Waymo Open dataset demonstrate the effectiveness of our approach.

%% file: srcs/7-acknowledgement.tex
\vspace{-1em}
\paragraph{Acknowledgments} The work is supported by the National Key R\&D Program of China (2020AAA0105200), Beijing Academy of Artificial Intelligence.

%% file: srcs/6-appendix.tex
\section*{Appendix}
\appendix

\section{More Dataset and Implementation Details}
\label{sec:more_details}

\subsection{Dataset Details}

\noindent\textbf{Waymo Open dataset~\cite{sun2019scalability}~} collects 1,000 2D video sequences and 3D point clouds at synchronized time steps. There are around 200 frames captured at 10Hz in each video. Raw data from 5 LiDARs (1 mid-range and 4 short-range) and 5 cameras (1 front and 4 sides) is provided. In experiments, 2D images from the `front' camera and 3D point clouds from the `top' LiDAR are utilized. The temporal information from videos is not used.
There are 3 object categories annotated with 2D bounding boxes in the dataset, i.e., `vehicles', `pedestrians', and `cyclists'.

\vspace{0.5em}
\noindent\textbf{Cityscapes dataset~\cite{Cordts2016Cityscapes}~} is a large-scale dataset for the autonomous driving scenario. The training and validation sets contain 2975 and 500 images, respectively. Each image is annotated with high quality pixel-level instance annotations. All images are of a resolution of 1024$\times$2048 pixels. Feature transferability experiments are conducted on the instance segmentation task, which involves 8 object categories.

\vspace{0.5em}
\noindent\textbf{PASCAL VOC dataset~\cite{everingham2015pascal}~} is consisted of the PASCAL VOC 2007 and 2012. Feature transferability experiments are conducted on the object detection task, which involves 20 object categories. Subsets trainval2007 and trainval2012 are used for training, and subset test2007 is used for validation. The training and validation sets contain 17k and 5k images, respectively.

\subsection{Implementation Details}
\label{sec:implementation_appendix}

\noindent\textbf{Training MoCo v2 on Waymo Open dataset~\cite{sun2019scalability}~}
As for the MoCo feature learning for segment label initialization, we extend the MoCo v2~\cite{chen2020improved} with 3D point clouds as the additional inputs.
Given the 3D segment $\varsigma_i$, PointNet~\cite{qi2017pointnet} is applied to extract a 1024-d 3D shape feature. Given the image $x$ and the 2D bounding box $b_i$, ResNet-50~\cite{he2016deep} is applied to extract a 2048-d image appearance feature from 2D image patch within bounding box $b_i$. We conduct the MoCo feature learning based on the concatenation of the 3D shape feature and 2D image appearance feature, which are further fed into a two-layer MLP with 128-d intermediate channels as in MoCo v2.

For the MoCo feature learning, augmentations for 2D image patches are exactly the same as in MoCo v2~\cite{chen2020improved}. Augmentations for 3D segments mainly follow PointNet~\cite{qi2017pointnet}, which consist of 1) random rotate along the up-axis with the angle in $[0, 2\pi]$, 2) horizontal random flip with probability of $0.5$, and 3) random dropout points with the probability uniformly sampling from $[0, 0.875]$, and pad to 1024 points by re-sampling from the kept points.

During the MoCo feature learning, all hyperparameters follow the MoCo v2 ~\cite{chen2020improved}. After that, k-means clustering~\cite{bahmani2012scalable} is performed on the 3072-d features, which is the concatenation of the 1024-d 3D shape feature and 2048-d 2D image appearance feature.

\vspace{0.5em}
\noindent\textbf{Training Segment Labeling Networks on Waymo Open dataset~\cite{sun2019scalability}~}
The segment labeling network $N$ takes both the 2D image and 3D point cloud as input. Given the 3D segment $\varsigma_i$, PointNet~\cite{qi2017pointnet} is applied to extract a 1024-d 3D shape feature. Given the image $x$ and the 2D bounding box $b_i$, Fast R-CNN~\cite{girshick2015fast} with ResNet-50~\cite{he2016deep} is applied to extract a 1024-d 2D image appearance feature. The concatenation of these two features is further fed into a linear classifier. The Batch Normalization (BN)~\cite{ioffe2015batch} layers are replaced with Synchronized BN~\cite{peng2018megdet}.
Models are trained on images of shorter side \{480, 512, 544, 576, 608, 640, 672, 704, 736, 768, 800\} pixels. In inference, images are resized so that the shorter side is 800 pixels.

In SGD training, 128 jittered segments (at a positive-negative ratio of 1:3) are sampled for each image. The networks are trained on 8 GPUs with 4 images per GPU for 6k iterations. The learning rate is initialized to 0.04 and is divided by 10 at the 3k and the 4k iterations. The weight decay and the momentum parameters are set as $10^{-4}$ and 0.9, respectively.

\vspace{0.5em}
\noindent\textbf{Training Object Detectors on Waymo Open dataset~\cite{sun2019scalability}~}
As for the object detector, we utilize FPN~\cite{lin2016feature} with ResNet-50~\cite{he2016deep} as the backbone. The Batch Normalization (BN)~\cite{ioffe2015batch} layers are replaced with Synchronized BN~\cite{peng2018megdet}.
The other choice of hyperparameters for Faster R-CNN~\cite{ren2015faster} follows the latest Detectron2~\cite{wu2019detectron2} code base, which is briefly presented here. Models are trained on images of shorter side \{480, 512, 544, 576, 608, 640, 672, 704, 736, 768, 800\} pixels. In inference, images are resized so that the shorter side is 800 pixels. Anchors are of 5 scales and 3 aspect ratios. 1k region proposals are generated at an NMS threshold of 0.7. During inference, detection results are derived at an NMS threshold of 0.3.

In SGD training, 256 anchor boxes (at a positive-negative ratio of 1:1) and 128 region proposals (at a positive-negative ratio of 1:3), are sampled for RPN and Fast R-CNN, respectively. The networks are trained on 8 GPUs with 4 images per GPU for 6k iterations. The learning rate is initialized to 0.04 and is divided by 10 at the 3k and the 4k iterations. The weight decay and the momentum parameters are set as $10^{-4}$ and 0.9, respectively.

\vspace{0.5em}
\noindent\textbf{Transfer learning on Cityscapes dataset~\cite{Cordts2016Cityscapes} and PASCAL VOC dataset~\cite{everingham2015pascal}~}
Mask R-CNN~\cite{he2017mask} and Faster R-CNN~\cite{ren2015faster} are utilized for instance segmentation and object detection, respectively. ResNet-50~\cite{he2016deep} with FPN~\cite{lin2016feature} is utilized as the backbone. For all experiments, the Batch Normalization (BN)~\cite{ioffe2015batch} layers are replaced with Synchronized BN~\cite{peng2018megdet}. The other choice of hyperparameters mainly follow \cite{he2019momentum}, which is briefly presented here.

For experiments on Cityscapes, models are trained on images of shorter sides \{800, 832, 864, 896, 928, 960, 992, 1024\} pixels, and tested on its original resolution of 1024$\times$2048 pixels. In SGD training, the networks are trained on 8 GPUs with 1 image per GPU for 24k iterations. The learning rate is initialized to 0.01 and is divided by 10 at the 18k iterations. 

For experiments on PASCAL VOC, models are trained on images of shorter sides \{480, 512, 544, 576, 608, 672, 704, 736, 768, 800\} pixels, and tested on images with a shorter side of 800 pixels. In SGD training, the networks are trained on 8 GPUs with 2 images per GPU for 54k iterations. The learning rate is initialized to 0.02 and is divided by 10 at the 36k and 48k iterations. 

\section{More Ablations}
\label{sec:more_ablations}

\setlength{\tabcolsep}{3pt}
\setlength{\doublerulesep}{2\arrayrulewidth}
\renewcommand{\arraystretch}{1.2}
\begin{table}[t]
    \centering
    \small
    \resizebox{0.99\linewidth}{!}{
        \begin{tabular}{c|ccc|ccc|ccc}
            \Xhline{2\arrayrulewidth}
            \multirow{2}{*}{\makecell[c]{foreground \\ threshold $\eta$}} & \multicolumn{3}{c|}{vehicle} & \multicolumn{3}{c|}{pedestrains} & \multicolumn{3}{c}{cyclist} \\
             & iter1 & iter2 & iter3 & iter1 & iter2 & iter3 & iter1 & iter2 & iter3 \\
            \hline
            0.999 & 23.0 & 24.9 & 26.2 & 11.5 & 12.2 & 15.0 & 0.0 & 1.0 & 0.0 \\
            0.99 & 26.3 & 26.9 & 27.6 & 17.1 & 21.8 & 23.1 & 2.3 & 2.3 & 1.7 \\
            \textbf{0.95} & 26.9 & 27.7 & 28.5 & 21.1 & 23.8 & 25.6 & 5.2 & 5.5 & 6.0 \\
            0.90 & 27.7 & 27.7 & 27.8 & 21.8 & 24.6 & 26.1 & 3.6 & 4.5 & 5.9 \\
            0.80 & 27.0 & 27.7 & 27.8 & 22.6 & 25.5 & 26.5 & 4.5 & 4.3 & 3.8 \\
            \Xhline{2\arrayrulewidth}
        \end{tabular}
    }
    \caption{Ablation on foreground threshold $\eta$ in the iterative segment labeling on Waymo Open validation. The default setting is highlighted.}
    \label{table:foreground_threshold}
    \vspace{-0.5em}
\end{table}

\noindent\textbf{Foreground Probability Threshold for Iterative Labeling~}
During the process of iterative segment labeling, a segment is labeled as background if its estimated foreground probabilities are less than $\eta$ for all clusters. As shown in Table~\ref{table:foreground_threshold}, our iterative segment labeling is robust to the choice of foreground threshold within a wide range.

\begin{figure*}[t]
    \centering
    \includegraphics[width=1.0\linewidth]{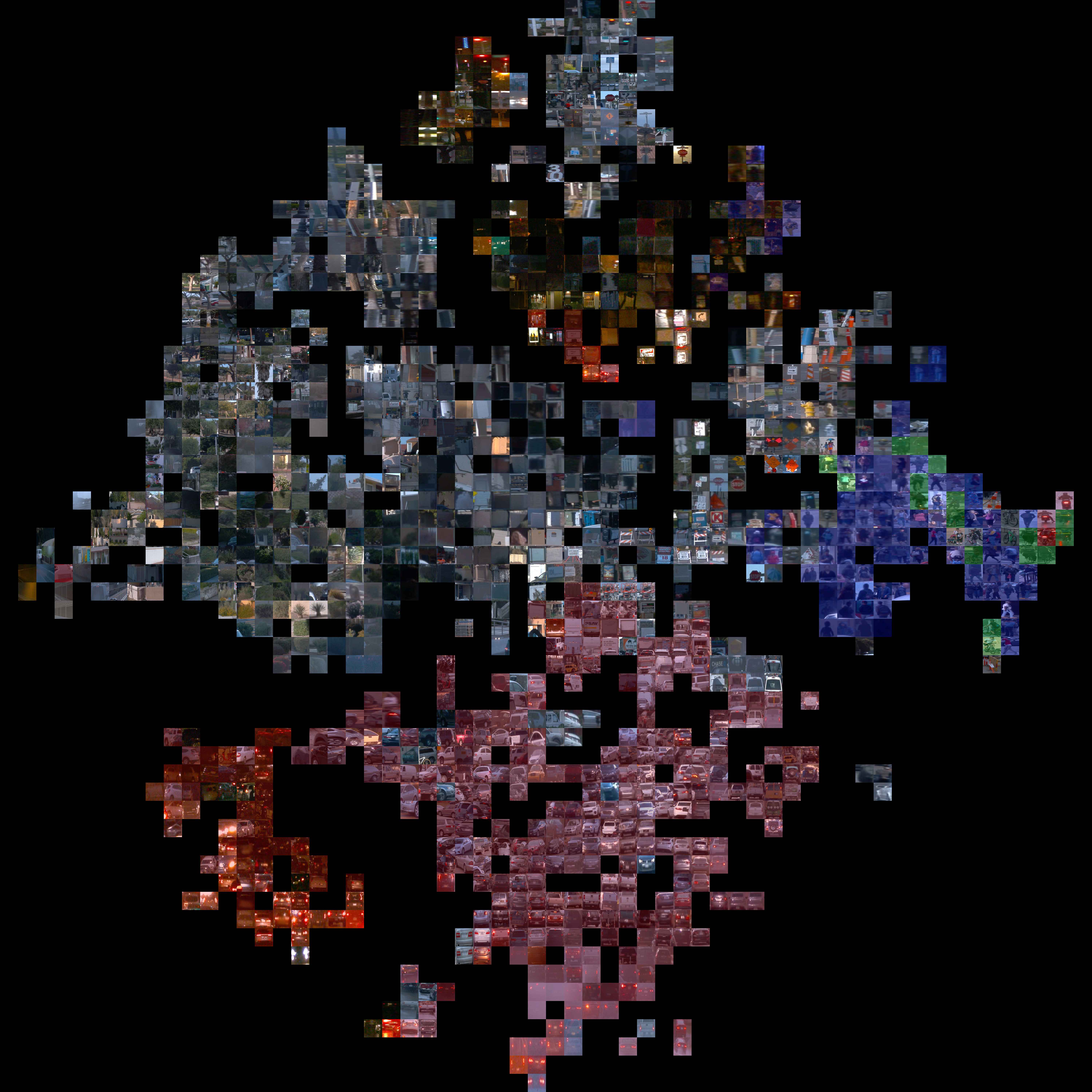}
    \caption{Visualization of t-SNE for features from the segment labeling network. Each embedding point in t-SNE corresponds to an candidate segment labeled as foreground by our method. Randomly sampled embedding points are visualized by the corresponding 2D image patches. Different color blendings represent difference categories: red for vehicle, blue for pedestrian, green for cyclist, and the remainings are unlabeled in the Waymo Open dataset.}
    \label{fig:image_tsne}
    \vspace{-1em}
\end{figure*}

\begin{figure*}[t]
    \centering
    \includegraphics[width=1.0\linewidth]{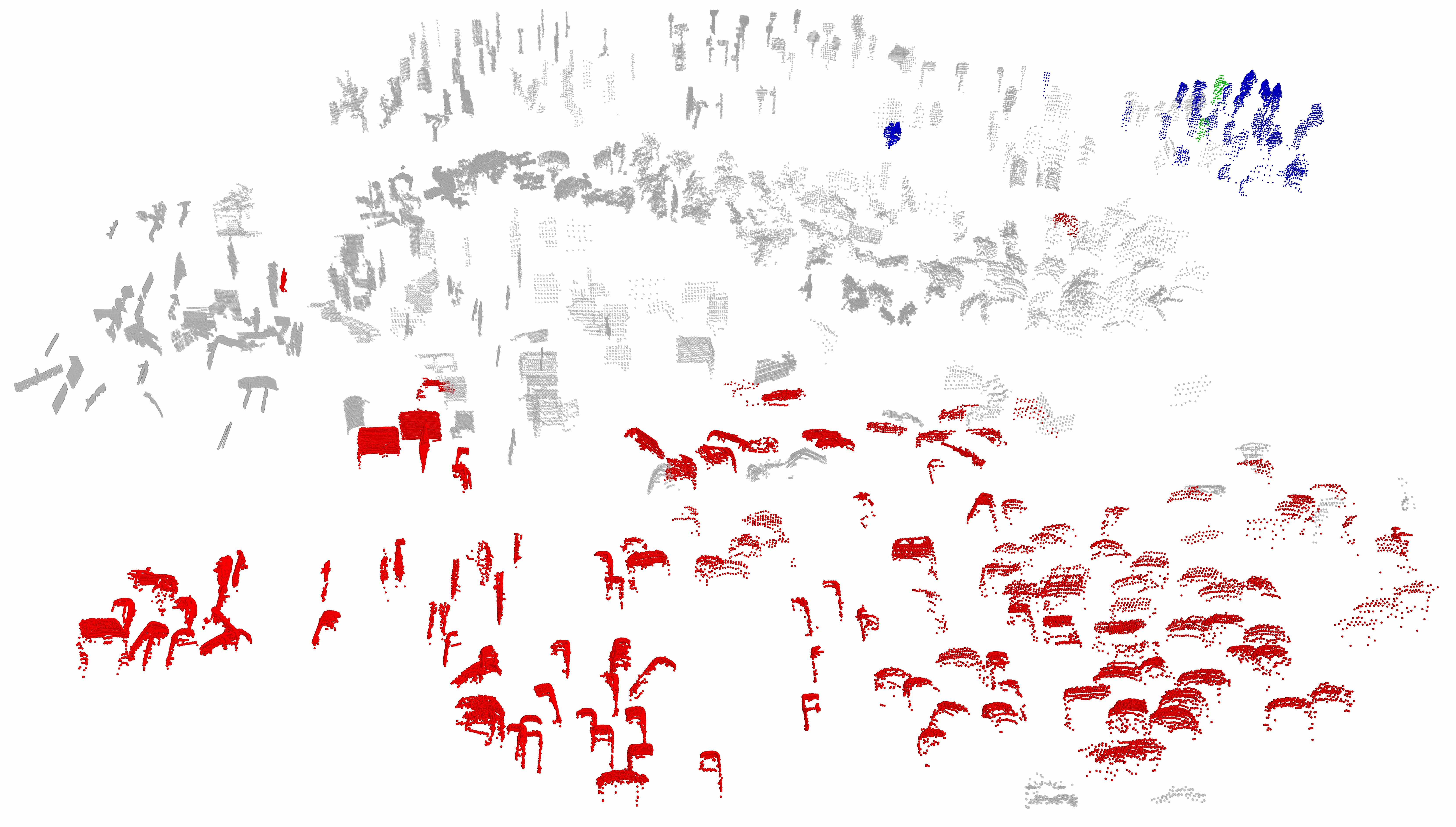}
    \caption{Visualization of t-SNE for features from the segment labeling network. Each embedding point in t-SNE corresponds to an candidate segment labeled as foreground by our method. Randomly sampled embedding points are visualized by the corresponding 3D LiDAR segments. Different colors represent difference categories: red for vehicle, blue for pedestrian, green for cyclist, and the remainings are unlabeled in Waymo Open dataset.}
    \label{fig:lidar_tsne}
    \vspace{-1em}
\end{figure*}

\vspace{0.5em}
\noindent\textbf{Visualization of Learned Feature Space~}
Since our method is unsupervised and has no access to semantic labels during training, it is interesting to see how well the learned features align with the semantics. Here, we show the t-SNE~\cite{maaten2008visualizing} for features learned by the segment labeling network after 10 rounds iterative labeling, which is the concatenation of the 1024-d 3D shape feature and the 1024-d 2D image appearance feature. Only candidate segments labeled as foreground by our method are shown in the embedding. Figure~\ref{fig:image_tsne} and Figure~\ref{fig:lidar_tsne} show the t-SNE visualizations, where each embedding point is represented by an image patch and a LiDAR segment, respectively.
The learned features could distinguish objects of different categories (e.g., car, pedestrian, and traffic cones) as well as the semantic subgroups of the same category (e.g., front, side, and rare views).

